\documentclass[final,5p,times,twocolumn,authoryear]{elsarticle}

\usepackage{amssymb}
\usepackage{booktabs}
\journal{arXiv}

\begin{document}

\begin{frontmatter}



\title{Multi-organ Self-supervised Contrastive Learning for Breast Lesion Segmentation}


\author[inst1]{Hugo Figueiras}  \ead{hfigueiras@lasige.di.fc.ul.pt}
\author[inst1]{Helena Aidos}   \ead{haidos@ciencias.ulisboa.pt}
\author[inst1]{Nuno Garcia}    \ead{ncgarcia@ciencias.ulisboa.pt}

\affiliation[inst1]{organization={LASIGE},%Department and Organization
            addressline={Faculdade de Ciências, Departamento de Informática, Edifício C6 Piso 3 - Sala 6.3.30, Universidade de Lisboa}, 
            city={Lisbon},
            postcode={1749-016}, 
            country={Portugal}}

\begin{abstract}
Self-supervised learning has proven to be an effective way to learn representations in domains where annotated labels are scarce, such as medical imaging. A widely adopted framework for this purpose is contrastive learning and it has been applied to different scenarios. This paper seeks to advance our understanding of the contrastive learning framework by exploring a novel perspective: employing multi-organ datasets for pre-training models tailored to specific organ-related target tasks.
More specifically, our target task is breast tumour segmentation in ultrasound images. The pre-training datasets include ultrasound images from other organs, such as the lungs and heart, and large datasets of natural images. Our results show that conventional contrastive learning pre-training improves performance compared to supervised baseline approaches. Furthermore, our pre-trained models achieve comparable performance when fine-tuned with only half of the available labelled data. Our findings also show the advantages of pre-training on diverse organ data for improving performance in the downstream task.
\end{abstract}




\end{frontmatter}


\section{Introduction}
\label{sec:intro}

%%       \citet{<label>} ==> Jones et al. (2015)
%%       \citep{<label>} ==> (Jones et al., 2015)

The performance of machine learning tasks is related to the amount of labelled images, but building such annotated datasets can be difficult. The issue is aggravated in projects involving medical image analysis, where professional annotations are frequently needed, and crowdsourcing is not a straightforward option. Labelling the data is usually the most time-consuming and arduous phase in any medical image analysis task, and numerous strategies have been put forth to alleviate this issue in data annotation. Self-supervised learning (SSL) methods~\citep{contextPrediction,inpainting,jigsaw,unsupervisedRotation} are a viable approach to tackle this problem since they offer a pre-training strategy that solely uses unlabeled data that generates an appropriate initialization for training downstream tasks with limited labelled data.

Until recently, self-supervised techniques have had great success for downstream analysis of both natural~\citep{VOC_challange,ImageNet_challange} and medical images~\citep{SSL_Cardiac_Segmentation,SSL_Medical_context_resotarion,SSL_medical_rubiks_cube}. This work emphasises contrastive learning~\citep{simclr,moco,simclr_V2,MoCov2,limitedannotations}. This popular self-supervised learning variation focuses on learning representations that minimize the distance between different views of the same concept and maximize the distance between different concepts, using a so-called contrastive loss~\citep{dimenstionality_reduction_loss,InfoNCE_Loss,simclr}. The neural networks trained to minimize this loss extract image representations that can be used for downstream tasks and give a good initialization that can be fine-tuned to a downstream task.

Most contrastive learning methods were developed for pre-training models using natural images and with the downstream task of image classification. In this paper, we study three of the most popular contrastive learning frameworks, SimCLR~\citep{simclr,simclr_V2}, MoCo~\citep{moco,MoCov2} and SimSiam~\citep{SimSiam} directly applied to medical images, specifically breast ultrasounds, for the downstream task of image segmentation.

As of 2020, breast cancer has become the most commonly occurring cancer in the world, with the highest incidence rate and second highest mortality rate, surpassing lung cancer~\citep{breast_cancer_staistics} in women. Early diagnosis of breast abnormalities, especially malignant tumours, is critical for treating and improving patient outcomes~\citep{benefits_breast_cancer_screening}. Although mammography is commonly used as the initial screening method, ultrasound is frequently utilized to evaluate palpable lumps, clarify unclear mammogram results, or assist in biopsies.
This is crucial for younger women with dense breast tissue, as ultrasound can distinguish benign from malignant lesions.

There are several reasons why breast ultrasound is highly recommended for specific situations. Firstly, it does not rely on radiation, making it a safer option for repeated use and for specific groups like pregnant women, unlike mammography or CT scans. Secondly, it offers real-time imaging, enabling physicians to assess structures dynamically. Lastly, it is usually less costly than other methods like MRI~\citep{cost_breast_cancer}.

However, there are still some problems regarding ultrasound segmentation. Segmenting ultrasound images can be difficult due to particular challenges. The speckle pattern can create noise, making it difficult to achieve accurate segmentation. Additionally, the contrast between lesions, particularly benign ones, and the surrounding breast tissue can be quite low, requiring a high level of expertise for precise interpretation. Structural variability, differences in anatomy, and even the type of ultrasound device used can result in significant variations in image appearances.

The main goal of this paper is to provide insights to guide the development of novel SSL algorithms for medical applications. We focus on the effect of different architectures of encoder-decoder networks for segmentation and the effect of using datasets adjacent to the end task, such as using the same modality but different organs. We'll analyze how pre-training with multi-organ ultrasound data can help address the challenges of ultrasound segmentation.
% By understanding the state-of-the-art methods, adapting them to the medical domain and comparing the results with current contrastive learning improvements, we aim to establish a strong foundation for advancing the field of medical image analysis and enabling the development of more accurate methods.
The key insights and contributions of this work are:
\begin{itemize}
  % \item The creation of an ultrasound dataset formed by subsets of heart, chest and breast ultrasounds datasets. %hf:talvez remover isto, apenas estamos a juntar datasets
  \item A simple implementation of SimCLR, MoCo and SimSiam for the downstream task of breast tumour segmentation on ultrasound images (BUS dataset) shows improvements over the fully supervised counterpart.
  \item We investigate whether pre-training with data from various organs and different datasets provides benefits compared to pre-training solely with images from the target organ and natural images. We conducted experiments using three ultrasound datasets and compiled a dataset comprising images from these sources and used a large natural image dataset. Our findings confirm the advantages of multi-organ pre-training.
  \item We analyze the impact of self-supervised pre-training when fine-tuning models with decreasing amounts of labels. It can be observed that at some point, fine-tuning with fewer labels can achieve as good performance as fine-tuning with all the available labels. This interesting result must be further investigated and generalized to other tasks and modalities.
\end{itemize}

\section{Method}
Self-supervised learning refers to the idea of building a supervised learning task from unlabeled data, \textit{i.e.}, using different views of the data or the data itself as supervision signals. A simple example is a system that learns to predict part of its input from other parts, \textit{e.g.} predict a frame of a video given the previous one, predict a word given the surrounding words, and so on.

One kind of self-supervised method is contrastive learning. The basic idea of contrastive learning is that two data points of the same class, the positive pairs, should have similar embeddings, while two data points from different classes, the negative pairs, should have dissimilar embeddings. Positive and negative pairs are usually constructed using data augmentation techniques --- altering an image through different transformations does not change its semantic meaning. Hence, by applying transformations to an image, one can generate new images that look like the original and still keep its properties. 

The model is trained to maximize the separation of negative pairs and minimize the distance in latent space between positive pairs, which is usually referred to as pre-training with a pretext task - the task that is not the final one and serves as a pretext to learn useful features. The next stage is referred to as fine-tuning or supervised training, where the model is trained with a tiny amount of labelled samples to solve the end task, commonly named the downstream task.

Two state-of-the-art contrastive learning methods are SimCLR~\citep{simclr,simclr_V2} and MoCo~\citep{moco,MoCov2}. Both methods are designed to learn from unlabeled data powerful representations, which can later be fine-tuned for specific tasks such as image classification or object detection. 
SimCLR explores with in-batch samples that are created from the same mini-batch of both positive and negative pairings. MoCo, on the other hand, stores negative training samples in a dynamic dictionary with a queue and a moving-averaged encoder. SimSiam~\citep{SimSiam} fits a subtype of contrastive learning called instance discrimination. These methods eliminate the need for negative pairs and still offer a competitive performance.

\begin{figure}
  \centering
  \centerline{\includegraphics[width=9cm]{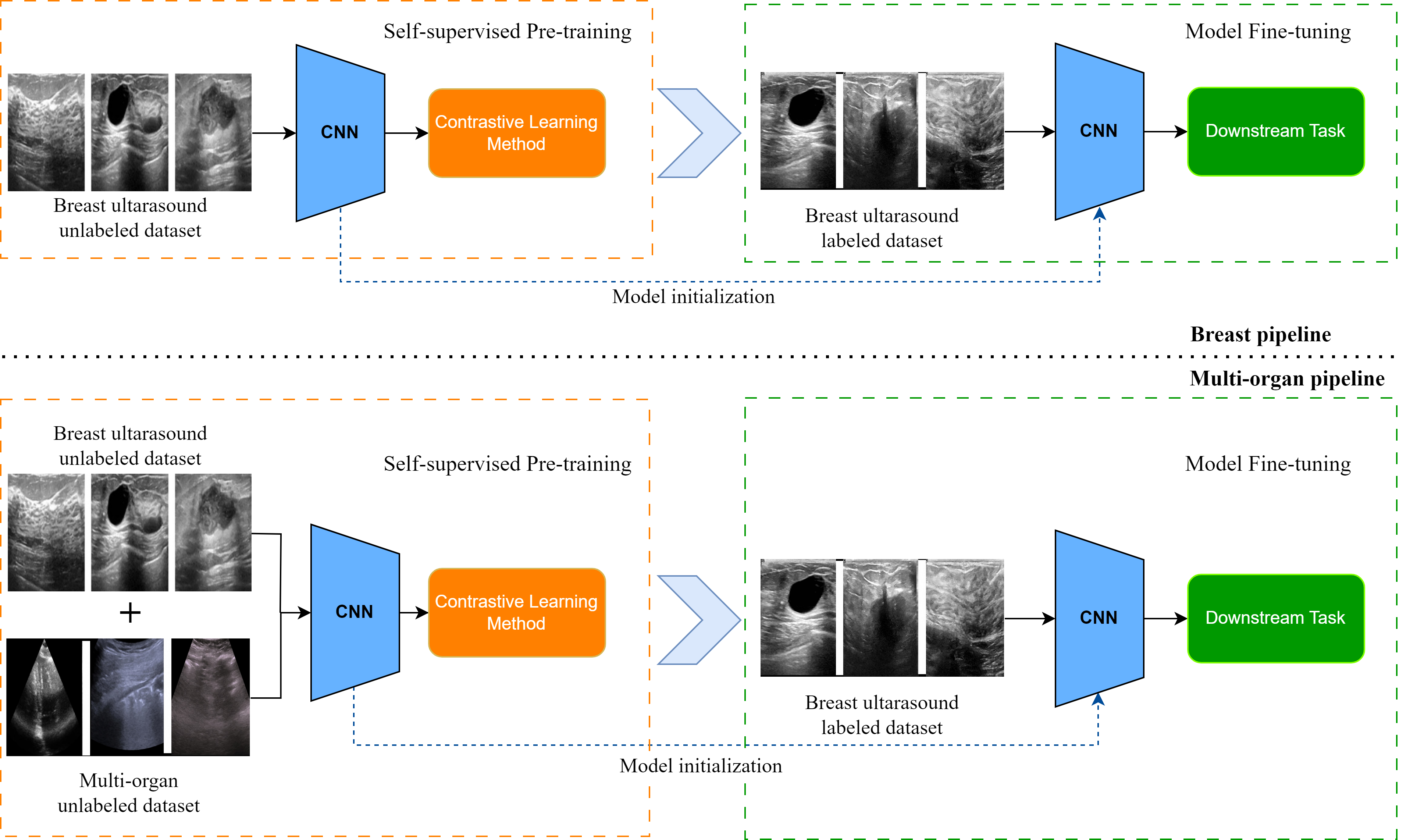}}
   \caption{Overview of the implemented method. The procedure starts with pre-training a model using a self-supervised learning method (SimCLR, MoCo or SimSiam) on an unlabeled dataset. The pre-trained weights are then used as the models' initialization when applied to a labelled dataset for the downstream task after the pre-training phase. In the breast pipeline, the breast ultrasound dataset is solely used for pre-training and in the multi-organ pipeline, datasets from different organs are complementary to the breast dataset.}
%  \vspace{2.0cm}
\label{fig:pipeline}
\end{figure}

With the developed methodology, we want to evaluate the impact of contrastive learning on ultrasound segmentation and study the effect of pre-training with ultrasounds from other organs different from the breast. Figure \ref{fig:pipeline} represents the overview of the implemented method, where each pipeline comprises two stages: the self-supervised pre-training and the fine-tuning. To begin, we pre-train a model using either the SimCLR, MoCo or SimSiam contrastive learning method on an unlabelled dataset. This process allows us to initialize the models' weights for the downstream task. Then, we fine-tune the model using a labelled dataset of the target organ we want to segment, which is the breast. The main difference between each pipeline lies in the datasets employed for pre-training. For the breast pipeline, we only utilize ultrasounds from the breast. However, we incorporate ultrasounds from the heart, lungs, and breast for the multi-organ pipeline.

\subsection{MoCo: Momentum Contrastive Learning}
MoCo~\citep{moco} views contrastive learning as a dictionary look-up task where the goal is to match a query to its appropriate key. It implements a dynamic dictionary as a queue with a momentum encoder. The dynamic dictionary contains a large number of keys, and one of the keys is a positive sample corresponding to the query, while all other keys are negative samples.

Each image is augmented, resulting in two augmented views of the same image $x_q$ and $x_k$. These augmented images $x_q$ and $x_k$ are then fed as input into two different encoders, the query encoder and the momentum encoder. The outputs from these encoders are normalized using L2-normalization, resulting in $q$ and $k_+$ that form a positive sample. MoCo trains the query encoder by maximizing the similarity between $q$ and $k_+$, which are views derived from the same image, while at the same time minimizing the similarity between $q$ and $k_i$, which are the negative samples. This similarity is enforced using a contrastive loss, namely the InfoNCE (Noise-Contrastive Estimation) loss defined as:
\begin{equation}
    \mathcal{L}_q = -\log\frac{\exp(q\cdot k_+/\tau)}{\sum_{i=0}^{K} \exp(q\cdot k_i/\tau) }. 
\end{equation}
New keys are added to the dynamic dictionary during each iteration, and the oldest batch is dequeued to eliminate stale embeddings. This makes the dictionary consistently reflect a sampled fraction of all data. Additionally, removing the oldest batch of mini-encoded keys can be advantageous since they are the most aged and, consequently, the least consistent with the newest ones. The query encoder is updated by backpropagation and the momentum encoder is updated using momentum. Given $\theta_k$ as the key encoder parameters, $\theta_q$ as the momentum encoder parameters, and $m$ as the momentum, the $\theta_k$ parameters are updated by:
\begin{equation}
    \theta_k = m\theta_k+(1-m)\theta_q.
\end{equation}

The set of transformations we apply to each image to generate different views are the same as those in MoCo V2: random horizontal flip, crop-and-resize, colour distortion, random grayscale, and Gaussian blur.

\subsection{SimCLR: Contrastive Learning}
SimCLR is a self-supervised learning framework introduced by \citet{simclr} in 2020. It leverages the concept of contrastive learning to learn representations from unlabeled data by maximizing agreement between two differently augmented views of the same data example using a contrastive loss in a hidden representation of neural networks~\citep{InfoNCE_Loss}. 
Given a randomly sampled mini-batch of images, each image $x$ is augmented twice using random horizontal flip, crop-and-resize, colour distortion, random grayscale, and Gaussian blur, resulting in two views of the same instance $\tilde{x_i}$ and $\tilde{x_j}$. To enable efficient training, it is crucial to have a good set of data augmentations since it directly influences how the latent space is organized and what patterns may be inferred from the data. 
The two views are then encoded using a base encoder $f(\cdot)$, usually a deep convolutional neural network, to extract the representation vectors $h_i$ and $h_j$ from the augmented data. These representations $h$ are mapped through the use of a multi-layer perceptron (MLP) projection head $g(\cdot)$ resulting in $z_i$  and $z_j$, to which the contrastive loss function is applied. In essence, this involves comparing the similarities between vectors.

This contrastive loss is the InfoNCE. By performing a softmax over the similarity values, this loss assesses the similarity between $z_i$ and $z_j$ relative to the similarity between $z_i$ and any other representation within the batch. This loss is defined as:
\begin{equation}
\ell_{ij} = -\log\frac{\exp(\textnormal{sim}(z_i,z_j)/\tau)}{\sum_{k=1}^{2N} \mathbb{1}_{[k\neq i]} \exp(\textnormal{sim}(z_i,z_k/\tau)},
\end{equation}
where \textnormal{sim(·,·)} is the cosine similarity between two vectors, and $\tau$ is a temperature scalar.

The model is trained by randomly sampling a batch of $N$ examples and defining the contrastive prediction task on pairs of augmented examples. 

\subsection{SimSiam: Siamese Representation Learning}

SimSiam~\citep{SimSiam} is a framework that proposes the use of Siamese networks to learn meaningful representations without using negative sample pairs, large batches or momentum encoders. The SimSiam method can be thought of as SimCLR without negative samples.

This framework takes in two different versions of an image, namely $x_1$ and $x_2$, which are then processed by a shared encoder network consisting of a backbone and a projection MLP in order to produce feature maps. The weights of the encoder are shared between the two views. A prediction MLP head $h$ is then applied to one of the views, which is then used to match it with the other view. Denoting the two output vectors as \(\rho_1 \triangleq h(f(x_1))\) and \(z_2 \triangleq f(x_2)\), its minimized their negative cosine similarity:
\begin{equation}
\label{eq:neg_cos_sim}
    \mathcal{D}(\rho_1,z_2) = - \frac{\rho_1}{\|\rho_1\|_2}\cdot\frac{\rho_2}{\|z_2\|_2},
\end{equation}
where $\|\cdot\|$ is $\ell_2$-norm. SimSiam uses a symmetric negative cosine similarity loss and therefore does not require any negative samples. This loss is defined as:
\begin{equation}
\label{eq:symmetrized_loss}
    \mathcal{L}=\frac{1}{2}\mathcal{D}(\rho_1,z_2)+\frac{1}{2}\mathcal{D}(\rho_2,z_1).
\end{equation}
This is defined for each image, and the total loss is averaged over all images. Its minimum possible value is $- 1$.

An important component to make this method work is using the stop-gradient (\texttt{stopgrad})  operation. This prevents the model from collapsing, and it is implemented by modifying Equation (\ref{eq:neg_cos_sim}) as:
\begin{equation}
    \mathcal{D}(\rho_1, \texttt{stopgrad}(z_2)).
\end{equation}
This means that $z_2$ is treated as a constant in this term. Equation (\ref{eq:symmetrized_loss}) is then implemented as:
\begin{equation}
    \mathcal{L}=\frac{1}{2}\mathcal{D}(\rho_1,\texttt{stopgrag}(z_2))+\frac{1}{2}\mathcal{D}(\rho_2,\texttt{stopgrag}(z_1))
\end{equation}

The results indicate that SimSiam performs better than other methods in ImageNet classification when pre-trained for 100 epochs, although the improvement with longer training is less significant. One of the significant advantages of the SimSiam methodology is that it uses fewer computational resources due to a smaller batch size.

\section{Experiments}
\subsection{Datasets}
\begin{figure*}[ht]
  \centering
  \includegraphics[width=\linewidth]{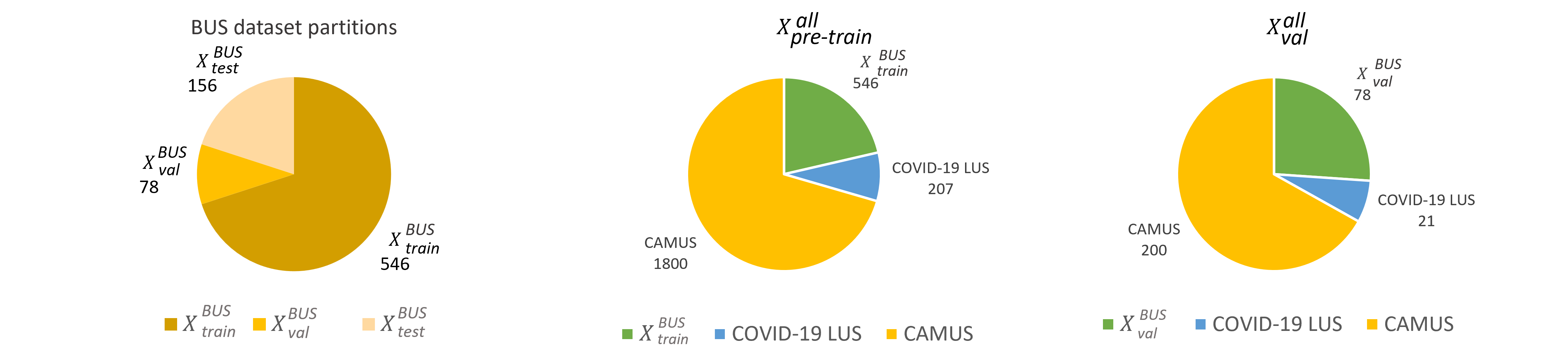}
   \caption{Visual representation of different dataset partitions. The BUS partitions are used in the fine-tuning of the models.}
%  \vspace{2.0cm}
\label{fig:graphs}
\end{figure*}
We utilize three well-known datasets of ultrasound images: the Breast Ultrasound Images Dataset (BUS)~\citep{DatasetBreastUltrasound}, with an average image size of 500$\times$500, the cardiac CAMUS dataset~\citep{CAMUS_Dataset}, composed by images of size 317$\times$317, and the COVID-19 Lung Ultrasound dataset (LUS)~\citep{born2021accelerating,born2021l2} containing images of various sizes.
We train and evaluate our models on a dataset composed of images from these three datasets. The combined dataset has 3008 images, of which 228 are from the COVID-19 LUS dataset, 2000 from CAMUS and 780 from BUS.

The CIFAR-10 dataset~\citep{CIFAR-10_dataset} is the fourth dataset used and is a multi-class classification dataset with ten object categories. It has 60,000 colour images of 32x32 pixels. The dataset is split into two subsets: a Training Set of 50000 images and a Validation Set of 10000 images.

The mini-ImageNet dataset, proposed by \citet{mini-ImageNet}, is used for evaluating few-shot learning. It contains 100 classes, with 600 samples per class. The dataset uses images from ImageNet~\citep{ImageNet_challange} and includes 60000 $84\times84$ colour images. We use a train partition of 48000 images and a validation partition of 12000 images for this work.

Since the BUS dataset is the one containing our target organ, the breast, we split it into a train $X^{BUS}_{train}$ partition to be used in the self-supervised pre-train and in supervised fine-tuning and validation $X^{BUS}_{val}$. We use $X^{BUS}_{val}$ for validation and test our models on a completely independent partition $X^{BUS}_{test}$. The size of these partitions are: $X^{BUS}_{train}=546$, $X^{BUS}_{val}=78$ and $X^{BUS}_{test}=156$.
The other datasets, CAMUS and COVID-19 LUS are split into a train and validation partition that will be used for the self-supervised pre-train and validation. By joining the partitions of the three datasets we get $X^{all}_{pre-train}$ with 2553 samples and $X^{all}_{val}$ with 299 samples, meaning that $X^{BUS}_{train} \subset X^{all}_{pre-train}$ and $X^{BUS}_{val} \subset X^{all}_{val}$. Figure \ref{fig:graphs} shows the different datasets partitions.
The BUS($\bigcirc$)+CIFAR-10 dataset contains 50546 images for pre-training and 10078 images for validation. The BUS($\bigcirc$)+mini-ImageNet contains 60546 and 12078 images for pre-training and validation, respectively.

\subsection{Pre-training protocol}
We investigate the effectiveness of self-supervised pre-training in ultrasound images using the U-Net architecture as our base network. The U-Net is augmented with a 2-layer MLP head with ReLU for self-supervised training and pre-trained end-to-end with $X^{all}_{pre-train}$ or $X^{BUS}_{train}$,  depending if using multi-organ or breast-only data, for training and $X_{val}$ sets for validation.
This setup will evaluate if the model improves with pre-training using images from the same modality but different organs besides the breast.

Following SimCLR~\citep{simclr} and MoCo v2~\citep{MoCov2}, two fully connected layers are used to map the output of the ResNet to a 128-dimensional embedding, which is used for contrastive learning and following SimSiam~\citep{SimSiam} this MLP has 3 layers. The U-Net is augmented with a 2-layer MLP (SimCLR and MoCo) and with a 3-layer MLP (SimSiam) head with ReLU for the %unsupervised training stage , pre-training
self-supervised training and pre-trained end-to-end. Contrastive learning pre-training uses $X^{all}_{pre-train}$ or $X^{BUS}_{train}$ for training, depending if using multi-organ or breast-only data,  and the corresponding $X_{val}$ sets for validation.
This setup will be used to evaluate if the model improves with pre-training using images from the same modality but different organs besides the breast.

To determine whether pre-training with multi-organ ultrasounds is beneficial due to its relation to the target task or simply because it provides additional data, we conducted experiments in which we pre-trained models using the CIFAR-10/mini-ImageNet dataset alone, as well as the CIFAR-10/mini-ImageNet dataset in combination with the BUS dataset.

We pre-train with different image sizes: $32\times32$ and $50\times50$. Experiments with $32\times32$ images are conducted on CIFAR-10, while mini-ImageNet is used for $50\times50$ images. This investigates the effect of image resolution on segmentation performance.

All experiments were run on an NVIDIA GeForce RTX 3060 GPU. During pre-training, a batch size of 512 is for SimCLR, 256 is used for MoCo, and 512 for SimSiam when working with images of size $32\times32$. When working with images of size $64\times64$ and $50\times50$, the batch size used for SimCLR, MoCo and SimSiam is 256, 64, and 512, respectively. We do not increase the batch size when pre-training with SimSiam since this makes the model collapse when using the BUS dataset or the Multi-organ dataset.

\subsection{Fine-tuning protocol}
The models are initialized with the weights obtained from the self-supervised pre-trained networks when applicable and fine-tuned in an end-to-end fashion. The pre-trained ResNet is used as a U-Net encoder with the same decoder as in the original architecture, meaning this decoder is not pre-trained. It is also used the ResNet-50 with the pre-trained decoder to study if the pre-trained decoder improves performance. In summary, the architectures used for fine-tuning are ResNet-50 with a U-Net decoder, ResNet-18 with a U-Net decoder, and U-Net and ResNet-50 with a pre-trained U-Net decoder. We focus on the results obtained with the ResNet-50 with a U-Net decoder and the vanilla U-Net.
The fine-tuning uses the $X^{BUS}_{train}$ dataset. 

In order to optimize each architecture, the images are resized to match the size used during pre-training. For example, if a model was pre-trained using images that were $32\times32$ in size, then for fine-tuning purposes, images of the same size ($32\times32$) will be used. This same principle applies to other image sizes as well. 

During fine-tuning, a learning rate of $1\times10^{-4}$ and a weight decay of $1\times10^{-6}$ are used. We are experimenting with different sizes of the $X^{BUS}_{train}$ subset, which includes 100\%, 50\%, 25\%, and 10\% of the available labelled data. This experiment will help us understand the impact of pre-training when using limited annotated data for fine-tuning. 

We run the fine-tuning experiments 10 times and report segmentation performance using the dice coefficient (DC) on the test set $X^{BUS}_{test}$.

\section{Quantitative Results}
\paragraph{\textbf{Encoder Pre-training}}
In Table \ref{tab_resnet_32} and Table \ref{tab_resnet}, we outline results from using the ResNet-50 architecture as the encoder. The first row details results from the fully supervised model, which combines a ResNet encoder with a U-Net decoder, trained end-to-end with the annotated data. The table's mid rows report the encoder model's performance pre-trained with SSL methods. Here, a U-Net decoder is later added, and the whole model is fine-tuned for the final task. The experiments of the bottom row of the table follow the same training strategy but use the multi-organ datasets for SSL pre-training. Each column presents the Dice Score regarding different amounts of annotated data used in training.
\begin{table*}[h]
\small
\caption{Task: breast ultrasound segmentation, measured by Dice Coefficient (DC) using a ResNet50 as the encoder of a U-Net model. The decoder is from the original U-net model and is not pre-trained. Except for the supervised baseline, each model is pre-trained on $X^{BUS}_{train} \equiv \bigcirc$ and $X^{all}_{pre\-- train} \equiv \bigtriangleup$ and fine-tuned using different amounts of the available labelled data from the $X^{BUS}_{train}$. All images used were of size $32\times32$. In the table below the results of the supervised baseline are presented for comparison purposes.}\label{tab_resnet_32}
    \centering
    \begin{tabular*}{\linewidth}{@{\extracolsep{\fill}} rccccc}
    \toprule
    \multicolumn{6}{c}{\textbf{U-Net model with ResNet50 encoder; only the encoder is pre-trained.}} \\
         Method & Dataset & DC (100\%) & DC (50\%) & DC (25\%) & DC (10\%)\\ \midrule
        Supervised & $\bigcirc$ & $0.587\pm0.039$ & $0.540\pm0.058$ & $\textbf{0.534}\pm0.028$ & $0.465\pm0.032$ \\ \midrule
        MoCo  &  & $0.558\pm0.038$ & $0.542\pm0.063$ & $0.521\pm0.031$ & $0.472\pm0.052$ \\ 
        SimCLR  & CIFAR-10 & $0.610\pm0.064$ & $0.553\pm0.057$ & $0.522\pm0.029$ & $0.477\pm0.045$\\
        SimSiam & & $0.629\pm0.030$ & $0.580\pm0.041$ & $0.451\pm0.129$ & $0.130\pm2.926$  \\ \midrule
        MoCo  &  & $0.561\pm0.025$ & $0.544\pm0.050$ & $0.505\pm0.051$ & $0.472\pm0.027$ \\ 
        SimCLR  & $\bigcirc$ & $0.590\pm0.046$ & $0.558\pm0.043$ & $0.530\pm0.046$ & $0.452\pm0.072$\\ 
        SimSiam  &  & $0.629\pm0.030$ & $0.567\pm0.035$ & $0.282\pm0.150$ & $0.163\pm0.104$\\ \midrule
        MoCo  &  & $0.602\pm0.033$ & $0.522\pm0.064$ & $\textbf{0.533}\pm0.050$ & $0.446\pm0.064$ \\ 
        SimCLR  & $\bigcirc$+CIFAR-10 & $0.600\pm0.033$ & $0.547\pm0.044$ & $0.507\pm0.036$ & $0.474\pm0.032$\\ 
        SimSiam  &  & $0.620\pm0.030$ & $0.597\pm0.046$ & $0.384\pm0.202$ & $0.403\pm0.186$\\ \midrule
        MoCo  &  & $0.628\pm0.030$ & $0.572\pm0.029$ & $0.498\pm0.066$ & $0.414\pm0.045$ \\ 
        SimCLR  & $\bigtriangleup$ & $0.592\pm0.058$ & $0.541\pm0.051$ & $0.509\pm0.039$ & $\textbf{0.496}\pm0.048$ \\
        SimSiam  & & $\textbf{0.638}\pm0.036$ & $\textbf{0.579}\pm0.036$ & $0.399\pm0.188$ & $0.404\pm0.150$ \\ \bottomrule
    \end{tabular*}
\end{table*}

The results of pre-training a ResNet-50 encoder using images of size $32\times32$ and then adding a U-Net decoder and fine-tuning it end-to-end are shown in Table \ref{tab_resnet_32}. It was observed that pre-training with multi-organ ($\bigtriangleup$) achieved the best results when fine-tuning with 100\%, 50\% of available labels using SimSiam and with 10\% using SimCLR. Additionally, MoCo pre-training with multi-organ data achieved the second-best results when fine-tuning with 100\% and 50\% of labels. These results surpassed the supervised baseline. 

However, when fine-tuning with 25\% of labels, the supervised baseline achieved the best results. It is worth mentioning that the supervised baseline only showed an improvement of 0.001 pp compared to MoCo pre-training with BUS ($\bigcirc$) + CIFAR-10 datasets and 0.004 pp compared to SimCLR pre-training with BUS ($\bigcirc$) dataset. This indicates that, although the supervised baseline was not surpassed, we can achieve similar performance with the referred pre-training. It has been observed that the models that include breast ultrasound data in pre-training achieve better results than the ones that only pre-train on natural images. For instance, the MoCo pre-trained on BUS ($\bigcirc$) model achieved better outcomes than any model pre-trained on CIFAR-10. This implies that it is preferable to pre-train a model with fewer samples but on a related task rather than pre-training a model with a larger dataset on natural images when fine-tuning with only 25\% of labels.

The mean dice coefficients obtained from the pre-trained models using SimCLR, MoCo, and SimSiam on the same dataset (Table \ref{tab_resnet_32}) are presented in Table \ref{tab_resnet_32_mean}. This table enables a global view analysis of the results regarding the pre-trained method and allows for a closer focus on the pre-training datasets. The latest findings reveal that pre-training with multi-organ data is advantageous when fine-tuning with 100\% and 50\% of labels, and the pre-trained models perform better than the supervised baseline. However, when the labels for fine-tuning are reduced to 25\% and 10\%, the supervised baseline achieves the highest accuracy. In such cases, CIFAR-10 and BUS ($\bigcirc$) + CIFAR-10 pre-training are the most effective pre-trained models.

The results of the study suggest that pre-training on multi-organ ($\bigtriangleup$) is advantageous. However, this advantage is maximized when fine-tuning with 100\% and 50\% of labels. Further testing is done to validate this hypothesis by increasing image resolution since segmentation performance improves with higher resolution.

\begin{table*}[h]
\small
\caption{Task: breast ultrasound segmentation, measured by Dice Coefficient (DC) using a ResNet50 as the encoder of a U-Net model. The decoder is from the original U-net model and is not pre-trained. Except for the supervised baseline, each model is pre-trained on $X^{BUS}_{train} \equiv \bigcirc$ and $X^{all}_{pre\-- train} \equiv \bigtriangleup$ and fine-tuned using different amounts of the available labelled data from the $X^{BUS}_{train}$. All images used were of size $64\times64$. In the table below the results of the supervised baseline are presented for comparison purposes.}\label{tab_resnet}
    \centering
    \begin{tabular*}{\linewidth}{@{\extracolsep{\fill}} rccccc}
    \toprule
    \multicolumn{6}{c}{\textbf{U-Net model with ResNet50 encoder; only the encoder is pre-trained.}} \\
         Method & Dataset & DC (100\%) & DC (50\%) & DC (25\%) & DC (10\%)\\ \midrule
        Supervised & $\bigcirc$ & $0.710\pm0.041$ & $0.629\pm0.075$ & $\textbf{0.630}\pm0.036$ & $0.531\pm0.061$ \\ \midrule
        MoCo  &  & $0.678\pm0.040$ & $0.627\pm0.041$ & $0.469\pm0.187$ & $0.320\pm0.163$ \\ 
        SimCLR  & mini-ImageNet & $0.693\pm0.066$ & $0.625\pm0.037$ & $0.611\pm0.040$ & $\textbf{0.561}\pm0.047$\\ 
        SimSiam  &  & $0.686\pm0.040$ & $0.627\pm0.051$ & $0.519\pm0.196$ & $0.313\pm0.163$\\ \midrule
        MoCo  &  & $0.695\pm0.025$ & $0.640\pm0.038$ & $0.541\pm0.083$ & $0.406\pm0.217$ \\ 
        SimCLR  & $\bigcirc$ & $0.691\pm0.050$ & $0.624\pm0.053$ & $\textbf{0.624}\pm0.035$ & $0.523\pm0.060$\\
        SimSiam  &  & $0.693\pm0.028$ & $0.624\pm0.032$ & $0.445\pm0.163$ & $0.408\pm0.108$\\ \midrule
        MoCo  &  & $0.693\pm0.040$ & $0.646\pm0.042$ & $0.435\pm0.175$ & $0.368\pm0.150$ \\ 
        SimCLR  & $\bigcirc$+mini-ImageNet & $0.694\pm0.026$ & $0.615\pm0.058$ & $0.615\pm0.037$ & $0.523\pm0.055$\\
        SimSiam  &  & $\textbf{0.714}\pm0.034$ & $0.638\pm0.038$ & $0.525\pm0.076$ & $0.466\pm0.183$\\ \midrule
        MoCo  &  & $0.686\pm0.031$ & $\textbf{0.658}\pm0.027$ & $0.453\pm0.182$ & $0.360\pm0.189$ \\ 
        SimCLR  & $\bigtriangleup$ & $0.694\pm0.053$ & $0.626\pm0.038$ & $0.608\pm0.054$ & $0.538\pm0.036$ \\ 
        SimSiam  & & $0.703\pm0.033$ & $0.653\pm0.042$ & $0.490\pm0.137$ & $0.305\pm0.180$ \\ \bottomrule
    \end{tabular*}
\end{table*}
It's worth noting that models pre-trained with SSL tend to perform better or at least equally well compared to their supervised counterparts when increasing image resolution. From Table \ref{tab_resnet}, it's evident that when fine-tuning with more than 25\% of labels, pre-trained models perform similarly regardless of the dataset used for pre-training. With 100\% of labels, only the SimSiam pre-trained on BUS($\bigcirc$) + mini-ImageNet model outperforms the supervised baseline, but all achieve competitive results and similar performance. With 50\% of labels, the best model is MoCo pre-trained on multi-organ ($\bigtriangleup$) data, which surpasses the supervised baseline. The other pre-trained models also show similar performance. With 25\% of labels, the best model overall is the supervised baseline, and the best pre-trained model is SimCLR pre-trained on the BUS ($\bigcirc$) dataset. With 10\% of labels, the best model is the SimCLR pre-trained on the mini-ImageNet dataset. Overall, SimSiam achieved the best results when fine-tuned with 100\% or 50\%, while with 25\% and 10\%, SimCLR achieved better results.

By examining Table \ref{tab_resnet_64_mean}, one can easily notice that the segmentation performance improves as the image resolution increases. Overall, the dice scores increase, and we can confirm that the models perform competitively when fine-tuned with 100\% and 50\%, respectively. Furthermore, the results show similar values for the different datasets on each fine-tuning partition.

To summarize, it appears that when using the ResNet-50 architecture with higher image resolution, multi-organ pre-training becomes less significant, unlike when pre-training and fine-tuning with $32\times32$ images. Although the pre-trained models performed similarly in general, using multi-organ ($\bigtriangleup$) for pre-training still yielded the best results. This improvement is more noticeable when pre-training with $32\times32$ images, and as we increase the image size to $64\times64$, the performance of all pre-trained models is comparable. Increasing image resolution also leads to higher dice coefficients. Overall, the SimSiam pre-training method performed the best and achieved the highest dice scores.

\paragraph{\textbf{Pre-training the encoder and the decoder}} In Table \ref{tab_unet_32} and Table \ref{tab_unet}, we outline results from using the U-Net architecture by pre-training the whole network, the encoder and decoder, as well as fine-tuning it. These tables follow the same format as the ones previously shown (Table \ref{tab_resnet_32} and Table \ref{tab_resnet}).
\begin{table*}[h]
\small
\caption{Task: breast ultrasound segmentation, measured by Dice Coefficient (DC) using the U-Net architecture. Both the encoder and decoder are pre-trained. Except for the supervised baseline, each model is pre-trained on $X^{BUS}_{train} \equiv \bigcirc$ and $X^{all}_{pre-train} \equiv \bigtriangleup$, and fine-tuned using different amounts of labelled data.  All images used were of size $32\times32$. In the table below, the results of the supervised baseline are presented for comparison purposes.}\label{tab_unet_32}
    \centering
    \begin{tabular*}{\linewidth}{@{\extracolsep{\fill}} rccccc}
    \toprule
    \multicolumn{6}{c}{\textbf{U-Net model with both encoder and decoder pre-trained using images of size $32\times32$.}} \\
         Method & Dataset & DC (100\%) & DC (50\%) & DC (25\%) & DC (10\%)\\ \midrule
        Supervised & $\bigcirc$ & $0.567\pm0.012$ & $0.544\pm0.014$ & $0.393\pm0.018$ & $0.198\pm0.014$ \\ \midrule
        MoCo  &  & $0.615\pm0.049$ & $0.548\pm0.027$ & $0.496\pm0.036$ & $0.461\pm0.053$ \\ 
        SimCLR  & CIFAR-10 & $0.506\pm0.020$ & $0.503\pm0.053$ & $0.484\pm0.057$ & $0.382\pm0.148$\\ 
        SimSiam  &  & $0.550\pm0.025$ & $0.509\pm0.021$ & $0.433\pm0.115$ & $0.382\pm0.079$\\    \midrule
        MoCo  &  & $0.510\pm0.030$ & $0.469\pm0.017$ & $0.401\pm0.047$ & $0.394\pm0.080$ \\ 
        SimCLR  & $\bigcirc$ & $0.523\pm0.021$ & $0.495\pm0.033$ & $0.477\pm0.024$ & $0.447\pm0.028$\\ 
        SimSiam  &  & $0.545\pm0.017$ & $0.510\pm0.024$ & $0.483\pm0.023$ & $0.426\pm0.037$\\    \midrule
        MoCo  &  & $\textbf{0.621}\pm0.040$ & $\textbf{0.604}\pm0.037$ & $\textbf{0.526}\pm0.030$ & $\textbf{0.486}\pm0.052$ \\ 
        SimCLR  & $\bigcirc$+CIFAR-10 & $0.468\pm0.152$ & $0.475\pm0.131$ & $0.430\pm0.145$ & $0.245\pm0.160$\\ 
        SimSiam  &  & $0.556\pm0.024$ & $0.524\pm0.031$ & $0.482\pm0.033$ & $0.430\pm0.035$\\    \midrule
        MoCo  &  & $0.563\pm0.027$ & $0.528\pm0.035$ & $0.489\pm0.025$ & $0.459\pm0.022$ \\ 
        SimCLR  & $\bigtriangleup$ & $0.558\pm0.012$ & $0.539\pm0.022$ & $0.521\pm0.013$ & $0.416\pm0.118$ \\ 
        SimSiam  & & $0.546\pm0.017$ & $0.487\pm0.037$ & $0.447\pm0.021$ & $0.409\pm0.032$ \\    \bottomrule
    \end{tabular*}
\end{table*}

In Table \ref{tab_unet_32}, we present the results of pre-training both the encoder and decoder of a standard U-Net model using images of size $32\times32$. Our experiments show that the best-performing model is the one pre-trained with MoCo on the BUS ($\bigcirc$) + CIFAR-10 dataset in all fine-tuning fractions. When fine-tuning using 100\% and 50\% of labels, the second-best method is the model pre-trained with MoCo on the CIFAR-10 dataset. Interestingly, unlike the results shown in Table \ref{tab_resnet_32}, when pre-training using the U-Net model with images of size $32\times32$, pre-training with a large general dataset seems to perform better than pre-training with the multi-organ dataset. We notice a significant difference in the dice scores of the MoCo pre-training on BUS ($\bigcirc$) + CIFAR-10 dataset with the multi-organ pre-training models. However, this difference is reduced when fine-tuning with 25\% and 10\% of labels, and the second-best models are now the ones trained on multi-organ data.

Based on the average dice scores obtained in each dataset, it is noticeable that fine-tuning pre-trained models with 100\% using the CIFAR-10 dataset and multi-organ ($\bigtriangleup$) dataset resulted in similar performance. The models pre-trained with BUS ($\bigcirc$) + CIFAR-10, which were the best models in Table \ref{tab_unet_32}, on average, ranked third best. On the other hand, the models pre-trained on BUS ($\bigcirc$) + CIFAR-10 achieved the best performance when fine-tuning with 50\%, while the models pre-trained on multi-organ data ($\bigtriangleup$) showed a similar performance. When fine-tuning with 25\% and 10\%, the advantage of multi-organ pre-training was more evident, with the models achieving the best results. In summary, although multi-organ pre-training did not yield the best results when fine-tuning with 100\% and 50\% of labels, it demonstrated competitive results. When fine-tuning with 25\% and 10\% of labels, the models achieved the best results. These findings further support the benefits of multi-organ pre-training.
\begin{table*}[h]
\small
\caption{Task: breast ultrasound segmentation, measured by Dice Coefficient (DC) using the U-Net architecture. Both the encoder and decoder are pre-trained. Except for the supervised baseline, each model is pre-trained on $X^{BUS}_{train} \equiv \bigcirc$ and $X^{all}_{pre-train} \equiv \bigtriangleup$, and fine-tuned using different amounts of labelled data. In the table below, the results of the supervised baseline are presented for comparison purposes.}\label{tab_unet}
    \centering
    \begin{tabular*}{\linewidth}{@{\extracolsep{\fill}} rccccc}
    \toprule
    \multicolumn{6}{c}{\textbf{U-Net model with both encoder and decoder pre-trained using images of size $50\times50$.}} \\
        Method & Dataset & DC (100\%) & DC (50\%) & DC (25\%) & DC (10\%) \\ \midrule
        % \multicolumn{2}{c}{Supervised Model}
        Supervised & $\bigcirc$ & $0.606\pm0.040$ & $0.574\pm0.017$ & $0.544\pm0.014$ & $0.505\pm0.031$ \\         \midrule
        MoCo  &  & $0.637\pm0.050$ & $0.591\pm0.036$ & $0.535\pm0.14$ & $0.484\pm0.045$ \\ 
        SimCLR  & mini-ImageNet & $0.594\pm0.022$ & $0.569\pm0.028$ & $0.465\pm0.154$ & $0.364\pm0.162$\\
        SimSiam  &  & $0.597\pm0.042$ & $0.559\pm0.040$ & $0.463\pm0.121$ & $0.449\pm0.040$\\    \midrule
        MoCo & & $0.701\pm0.035$ & $0.687\pm0.057$ & $0.697\pm0.065$ & $0.672\pm0.074$\\ 
        SimCLR & $\bigcirc$ & $0.595\pm0.097$ & $0.581\pm0.015$ & $0.582\pm0.015$ & $0.568\pm0.010$ \\ 
        SimSiam &  & $0.573\pm0.012$ & $0.535\pm0.034$ & $0.502\pm0.015$ & $0.444\pm0.027$ \\   \midrule
        MoCo  &  & $0.617\pm0.044$ & $0.588\pm0.045$ & $0.539\pm0.040$ & $0.482\pm0.035$ \\ 
        SimCLR  & $\bigcirc$+mini-ImageNet & $0.599\pm0.032$ & $0.532\pm0.178$ & $0.393\pm0.219$ & $0.264\pm0.233$\\
        SimSiam  &  & $0.693\pm0.038$ & $0.587\pm0.039$ & $0.544\pm0.028$ & $0.506\pm0.032$\\    \midrule
        MoCo  &  & $\textbf{0.723}\pm0.032$ & $\textbf{0.720}\pm0.021$ & $\textbf{0.714}\pm0.029$ & $\textbf{0.688}\pm0.041$ \\ 
        SimCLR  & $\bigtriangleup$ & $0.647\pm0.036$ & $0.645\pm0.017$ & $0.637\pm0.024$ & $0.611\pm0.040$ \\ 
        SimSiam &  & $0.573\pm0.012$ & $0.520\pm0.037$ & $0.468\pm0.021$ & $0.440\pm0.036$ \\ 
        \bottomrule
    \end{tabular*}
\end{table*}

Table \ref{tab_unet} shows results for pre-training both the encoder and decoder of a vanilla U-Net. Both SSL models, MoCo and SimCLR, outperformed the supervised U-Net baseline with a greater margin when trained with multi-organ ($\bigtriangleup$) data.

The models trained with ultrasounds from different organs ($\bigtriangleup$) achieved a higher DC than the ones trained only with breast ultrasounds ($\bigtriangleup$ vs $\bigcirc$) and trained with mini-ImageNet, showing that learning general features from other organs is beneficial. 
Moreover, these results also show that pre-trained models achieved similar performance when fine-tuned using 100\% and 50\% of available labels. This is a great advantage for projects in the medical domain where annotated data is scarce --- it is still possible to achieve good results even with few labels. In this setup, MoCo was the best pre-training method, followed by SimCLR.

Based on the mean dice scores presented in Table \ref{tab_unet_mean}, one can observe an increase in the overall dice scores when the image resolution is increased. Additionally, it is evident that the multi-organ ($\bigtriangleup$) pre-training strategy is the most effective when fine-tuning across all labelled fractions. Notably, when using the U-Net architecture and pre-training with images of size $50\times50$, combining breast ultrasound data with natural images in pre-training appears to result in worse outcomes.

When comparing Table \ref{tab_unet_32} with Table \ref{tab_unet}, it becomes apparent that increasing the image resolution results in better model performance. Additionally, the model pre-trained on multi-organ ($\bigtriangleup$) data outperformed the ones pre-trained on general data such as CIFAR-10 when the image size was increased. Even after increasing the image resolution, the models pre-trained on general data (CIFAR-10 and mini-ImageNet) maintained a similar level of performance. However, the models pre-trained on multi-organ ($\bigtriangleup$) data showed a significant improvement in their performance, becoming the best-performing models. It is worth noting that the benefit of multi-organ pre-training is most noticeable when pre-training and fine-tuning with an image size of $50\times50$.

\section{Qualitative Results}
This section evaluates the segmentation masks produced by the model and focuses on their practical applicability. In the real world, physicians who are experts in the field will use these models to interpret the results. Even if the segmentation isn't perfect, physicians can manually segment the missing parts from the mask. What's important for them is to know where the lesion is located and get a general segmentation of the tumour. Physicians can also observe and classify the tumour as either benign or malignant, which is another critical factor. Therefore, the following analysis focuses on how models can segment benign and malignant lesions to enable physicians to easily classify the type of lesion they encounter. Benign lesions have a more circular shape, while malignant lesions have a more irregular shape. It's vital for the model to segment these irregularities to differentiate between these types of lesions. So, instead of only focusing on correctly segmented pixels, the main focus should be on how the models can segment the different shapes of the different types of lesions. You can find the figures mentioned below in \ref{sec:masks}.

\paragraph{\textbf{Encoder Pre-training}} Figure \ref{fig:resnet32_comp} represents the mask outputs of some pre-trained models from Table \ref{tab_resnet_32}. This figure contains masks from the best pre-trained multi-organ model, and we compare them to the second best model, which, in this case, were the pre-trained models on CIFAR-10 and on the BUS dataset. In the experiment, we compared the mask outputs of several pre-trained models from Table \ref{tab_resnet_32}, as shown in Figure \ref{fig:resnet32_comp}. The figure includes masks from the best pre-trained multi-organ model and the second-best model, which were the pre-trained models on CIFAR-10 and on the BUS dataset. 
Upon analyzing Figure \ref{fig:resnet32_comp}, we observed that all the models were able to segment benign tumours effectively. However, model (c) lacked more prediction than the other models. Additionally, the multi-organ model (d) showed some difficulties in segmenting regular circular shapes, but it was the best at capturing the irregular shapes of benign tumours. 
It is important to note that there is still much room for improvement, and the purpose of this experiment was to compare the results of the experimented pre-trained models and not with state-of-the-art methods.

Figure \ref{fig:resnet64_comp} shows the mask outputs of three models from Table \ref{tab_resnet}: the best multi-organ pre-trained model, the best model, the model pre-trained on BUS + mini-ImageNet using SimSiam, and the supervised baseline for comparison. 
The segmentation of benign tumours remains consistent across all models. However, when segmenting malignant lesions, increasing the resolution tends to capture more irregular shapes, resulting in competitive mask predictions. This finding is consistent with the analysis presented in Table \ref{tab_resnet}.

\paragraph{\textbf{Pre-training the encoder and the decoder}} Analysing now the pre-trained U-Nets, Figure \ref{fig:unet32_comp} shows mask outputs of some pre-trained from Table \ref{tab_unet_32}. It appears that the segmentation of benign lesions is satisfactory, but the segmentation of malignant lesions is proving to be difficult. The models seem to be smoothing out the irregular shapes of the lesions, which is not ideal. In general, there is no model that stands out as particularly effective, which is consistent with the findings presented in Table \ref{tab_unet_32}.

The figure presented as Figure \ref{fig:unet50_comp} displays the output masks of various pre-trained models from Table \ref{tab_unet}. As we can observe, increasing the image resolution leads to more details in the segmentation of malignant lesions, while benign segmentation still yields good results. The pre-trained model (d), which was trained on multi-organ data using MoCo, provides better results than the one pre-trained on BUS (c). The supervised baseline model (b) also shows good segmentation masks, but the multi-organ pre-trained model (d) tends to capture more irregular shapes.

\section{Related Work}
%Contrastive learning methods were created on the assumption that transformations applied to an image do not alter its semantic meaning. %ncg: estava don't, mas não se deve usar abreviações
%Therefore, different augmentations derived from the same image form a positive pair, %ncg: repara o salto que dás aqui: estás a falar de positive pair sem antes ter explicado como funciona e para que serve um positive pair. Não é fácil encaixar tudo, e muitas vezes é impossivel mesmo.
%while the other images and their augmentations form a %negative pair concerning the current instance. These methods employ a contrastive loss to enforce augmentations from positive pairs to be similar while at the same time decreasing this similarity from negative pairs. 
\paragraph{\textbf{Self-supervised learning in medical imaging}}
When dealing with medical imaging, obtaining large labelled datasets is challenging. This is because it requires domain-specific experts to accurately label the images, and this labelling process can be uncertain due to natural disagreements on how to label images correctly. Self-supervised learning methods for medical images have recently gained popularity due to their competitive performance and capacity to learn from a small number of annotations. Some of these methods try to incorporate domain knowledge to enhance the learning process. \citet{limitedannotations} showed the effectiveness of global and local contexts to learn important latent features; \citet{SSL_Cardiac_Segmentation} train models in a self-supervised manner by predicting anatomical positions; \citet{SSL_Medical_context_resotarion} propose a novel self-supervised learning strategy based on context restoration to better exploit unlabeled images; \citet{SSL_medical_rubiks_cube} pre-train 3D neural networks using cube rearrangement and cube rotation, which enforce networks to learn translational and rotational invariant features from raw 3D data. Other directions include self-paced learning~\citep{self-paced-contrastive-learning}, uncertainty estimation~\citep{uncertainty-estimation}, domain adaptation~\citep{domain-adaptation}, etc.

\paragraph{\textbf{Breast ultrasound segmentation}}
Deep learning-based methods significantly improved the accuracy of breast ultrasound segmentation. Convolutional Neural Networks (CNNs) proved to be highly effective in learning hierarchical features directly from the raw data. In medical images, the structural information among neighbouring regions is important and CNNs were designed to better utilize the spatial information, hence the performance improvement. One popular CNN architecture that produces state-of-the-art results in breast ultrasound segmentation is the U-Net~\citep{U-NET} and its variants~\citep{U-Netanditsvariants}. \citet{Deep_learning_method_BUS_Segmentation} use the U-Net and data augmentations to create a fully automatic breast ultrasound pipeline. \citet{UNeXt} proposed UNeXt, a variant of the U-Net with tokenized multilayer perceptron (MLP) blocks to reduce the number of parameters and computational complexity while also improving segmentation performance. \citet{SK-U-Net} developed a selective kernel (SK) U-Net CNN to adjust the network’s receptive field using an attention mechanism and fuse feature maps extracted with dilated and conventional convolutions. Regarding self-supervised learning, some studies propose their developed method and use the U-Net to evaluate its performance on breast ultrasound segmentation~\citep{Breast_lesion_segmentation_UL_limited_data,CR-SSL,MPS-AMS}.

\section{Conclusion}
In this paper, we study the performance of popular contrastive learning frameworks applied to ultrasound medical images for the downstream task of breast lesion segmentation. Our research underlines the advantages of leveraging SSL models for medical imaging, mainly when applied to the U-Net architecture. The MoCo, SimCLR and SimSiam models, in particular, consistently outperformed or achieved similar performance of traditional supervised baselines, offering a promising direction for future investigations.

The primary objective of this paper was to explore a new concept of pre-training that could be advantageous. The results indicate that the performance of pre-trained models is similar when fine-tuned with only half of the available labels or even fewer in some cases. 

The key insight of this paper is that pre-training using images from different organs can complement pre-training with images containing only the target organ. This reinforces the notion that utilizing generalized features from various organs can significantly improve model accuracy. Although the benefit of pre-training with a multi-organ dataset can be setup dependent, being more or less relevant regarding image size and architecture in use.

Regarding the segmented masks, it is clear that higher image resolution leads to more detailed predictions for segmented masks. Additionally, our findings demonstrate that multi-organ pre-training is effective in capturing the irregular shapes of lesions, resulting in improved segmentation for malignant tumours. Furthermore, our models provide good results for benign tumour segmentation across the board.
After pre-training with multiple organs, the benefits are evident. 

Furthermore, our findings show a notable consistency in the performance of pre-trained models, especially when fine-tuned with different amounts of available labels. The pre-trained models consistently outperform the supervised baseline. This observation is particularly salient for medical imaging domains where annotated datasets can be sparse, suggesting that robust results can still be attained with a restricted label dataset.

\section*{Acknowledgement}
This work was supported by Fundação para a Ciência e a Tecnologia (FCT) under project EXPL/CCI-COM/0656/2021 and the LASIGE research unit, ref. UIDB/00408/2020 and UIDP/00408/2020.

\appendix

\section{Avarage Dice Coefficients of the pre-trained models}
\label{sec:mean_dc}
This section presents the average dice coefficients of the pre-trained models on each dataset. This provides a comprehensive view of the results of the pre-trained method and facilitates a detailed analysis of the pre-training datasets.

\begin{table*}[h]
\small
\caption{Mean Dice Coefficients (DC) of the pre-trained ResNet-50 models on each dataset (Table \ref{tab_resnet_32}) using images of size $32\times32$. In the table below, the results of the supervised baseline are presented for comparison purposes}\label{tab_resnet_32_mean}
    \centering
    \begin{tabular*}{\linewidth}{@{\extracolsep{\fill}} rccccc}
    \toprule
    \multicolumn{5}{c}{\textbf{U-Net model with ResNet50 encoder - Mean Dice Coefficients (DC) on each dataset}} \\
         Dataset & Mean DC (100\%) & Mean DC (50\%) & Mean DC (25\%) & Mean DC (10\%)\\ \midrule
        $\bigcirc$-Supervised & $0.587\pm0.039$ & $0.540\pm0.058$ & $\textbf{0.534}\pm0.028$ & $0.465\pm0.032$ \\ \midrule 
        CIFAR-10 & $0.599$ & $0.558$ & $\textbf{0.498}$ & $0.360$\\  \midrule
        $\bigcirc$ & $0.593$ & $0.556$ & $0.439$ & $0.362$\\ \midrule
        $\bigcirc$+CIFAR-10 & $0.607$ & $0.555$ & $0.475$ & $\textbf{0.441}$\\  \midrule
        $\bigtriangleup$ & $\textbf{0.619}$ & $\textbf{0.564}$ & $0.469$ & $0.438$ \\    \bottomrule
    \end{tabular*}
\end{table*}

\begin{table*}[h]
\small
\caption{Mean Dice Coefficients (DC) of the pre-trained ResNet-50 models on each dataset (Table \ref{tab_resnet}) using images of size $64\times64$. In the table below, the results of the supervised baseline are presented for comparison purposes}\label{tab_resnet_64_mean}
    \centering
    \begin{tabular*}{\linewidth}{@{\extracolsep{\fill}} rccccc}
    \toprule
    \multicolumn{5}{c}{\textbf{U-Net model with ResNet50 encoder - Mean Dice Coefficients (DC) on each dataset}} \\
         Dataset & Mean DC (100\%) & Mean DC (50\%) & Mean DC (25\%) & Mean DC (10\%)\\ \midrule
        $\bigcirc$-Supervised & $\textbf{0.710}\pm0.041$ & $0.629\pm0.075$ & $0.630\pm0.036$ & $\textbf{0.531}\pm0.061$ \\ \midrule 
        mini-ImageNet & $0.686$ & $0.626$ & $0.537$ & $0.446$\\  \midrule
        $\bigcirc$ & $0.693$ & $0.629$ & $\textbf{0.537}$ & $0.446$\\ \midrule
        $\bigcirc$+mini-ImageNet & $\textbf{0.700}$ & $0.633$ & $0.525$ & $\textbf{0.452}$\\  \midrule
        $\bigtriangleup$ & $0.694$ & $\textbf{0.646}$ & $0.517$ & $0.401$ \\    \bottomrule
    \end{tabular*}
\end{table*}

\begin{table*}[h]
\small
\caption{Mean Dice Coefficients (DC) of the pre-trained U-Net models on each dataset (Table \ref{tab_unet_32}) using images of size $32\times32$. In the table below, the results of the supervised baseline are presented for comparison purposes}\label{tab_unet_32_mean}
    \centering
    \begin{tabular*}{\linewidth}{@{\extracolsep{\fill}} rccccc}
    \toprule
    \multicolumn{5}{c}{\textbf{U-Net model with pre-trained encoder and decoder - Mean Dice Coefficients (DC) on each dataset}} \\
         Dataset & Mean DC (100\%) & Mean DC (50\%) & Mean DC (25\%) & Mean DC (10\%)\\ \midrule
        $\bigcirc$-Supervised & $\textbf{0.567}\pm0.012$ & $0.544\pm0.014$ & $0.393\pm0.018$ & $0.198\pm0.014$ \\ \midrule 
        CIFAR-10 & $\textbf{0.557}$ & $0.520$ & $0.471$ & $0.408$\\  \midrule
        $\bigcirc$ & $0.526$ & $0.491$ & $0.454$ & $0.422$\\ \midrule
        $\bigcirc$+CIFAR-10 & $0.548$ & $\textbf{0.534}$ & $0.479$ & $0.387$\\  \midrule
        $\bigtriangleup$ & $0.556$ & $0.518$ & $\textbf{0.486}$ & $\textbf{0.428}$ \\    \bottomrule
    \end{tabular*}
\end{table*}

\begin{table*}[h]
\small
\caption{Mean Dice Coefficients (DC) of the pre-trained U-Net models on each dataset (Table \ref{tab_unet}) using images of size $50\times50$. In the table below, the results of the supervised baseline are presented for comparison purposes}\label{tab_unet_mean}
    \centering
    \begin{tabular*}{\linewidth}{@{\extracolsep{\fill}} rccccc}
    \toprule
    \multicolumn{5}{c}{\textbf{U-Net model with pre-trained encoder and decoder - Mean Dice Coefficients (DC) on each dataset}} \\
         Dataset & Mean DC (100\%) & Mean DC (50\%) & Mean DC (25\%) & Mean DC (10\%)\\ \midrule
        $\bigcirc$-Supervised & $0.606\pm0.040$ & $0.574\pm0.017$ & $0.544\pm0.014$ & $0.505\pm0.031$ \\ \midrule 
        mini-ImageNet & $0.609$ & $0.573$ & $0.488$ & $0.432$\\  \midrule
        $\bigcirc$ & $0.623$ & $0.601$ & $0.594$ & $0.561$\\ \midrule
        $\bigcirc$+mini-ImageNet & $0.618$ & $0.569$ & $0.492$ & $0.417$\\  \midrule
        $\bigtriangleup$ & $\textbf{0.648}$ & $\textbf{0.628}$ & $\textbf{0.606}$ & $\textbf{0.580}$ \\    \bottomrule
    \end{tabular*}
\end{table*}

\section{Masks predictions of the pre-trained models}
\label{sec:masks}
In this appendix, we have included the mask predictions of several samples in the $X^{BUS}_{test}$ dataset. Each figure showcases the mask predictions of two models: the best multi-organ pre-trained model and the best or second-best model, if the multi-organ model is the best one, from the same table. Additionally, we included the mask predictions of the supervised baseline for comparison purposes. The first column of every figure displays benign lesions, while the second column shows malignant lesions.

\begin{figure}
  \centering
  \centerline{\includegraphics[width=9cm]{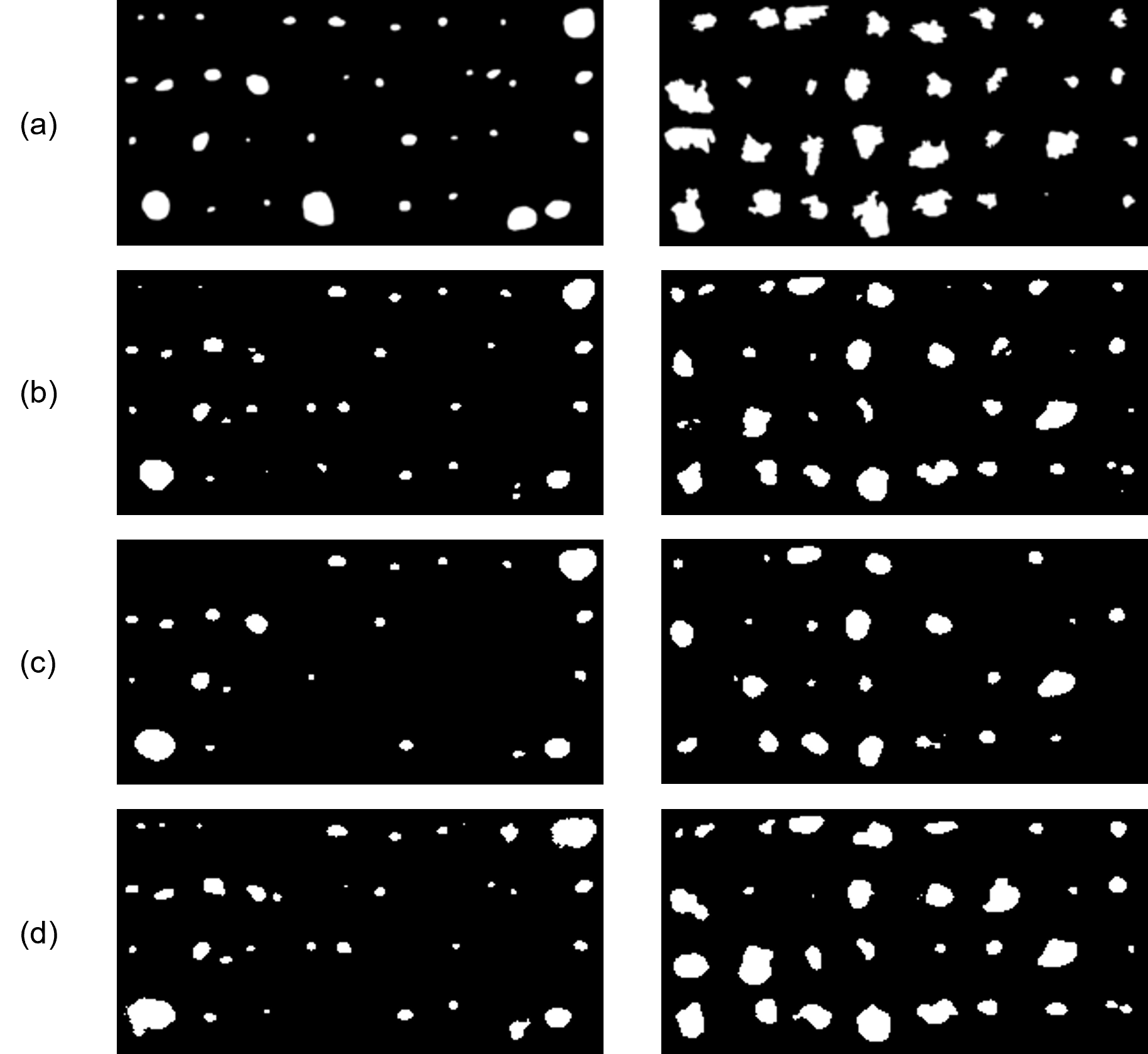}}
   \caption{Generated masks of the pre-trained ResNet-50 backbones, pre-trained and fine-tuned using $32\times32$ images. The first column shows the masks of benign tumours, and the second column shows the masks of malignant tumours. (a) Ground truth; (b) SimSiam – CIFAR-10; (c) SimSiam - BUS ($\bigcirc$); (d) SimSiam - Multi-organ ($\bigtriangleup$).
}
%  \vspace{2.0cm}
\label{fig:resnet32_comp}
\end{figure}

\begin{figure}
  \centering
  \centerline{\includegraphics[width=9cm]{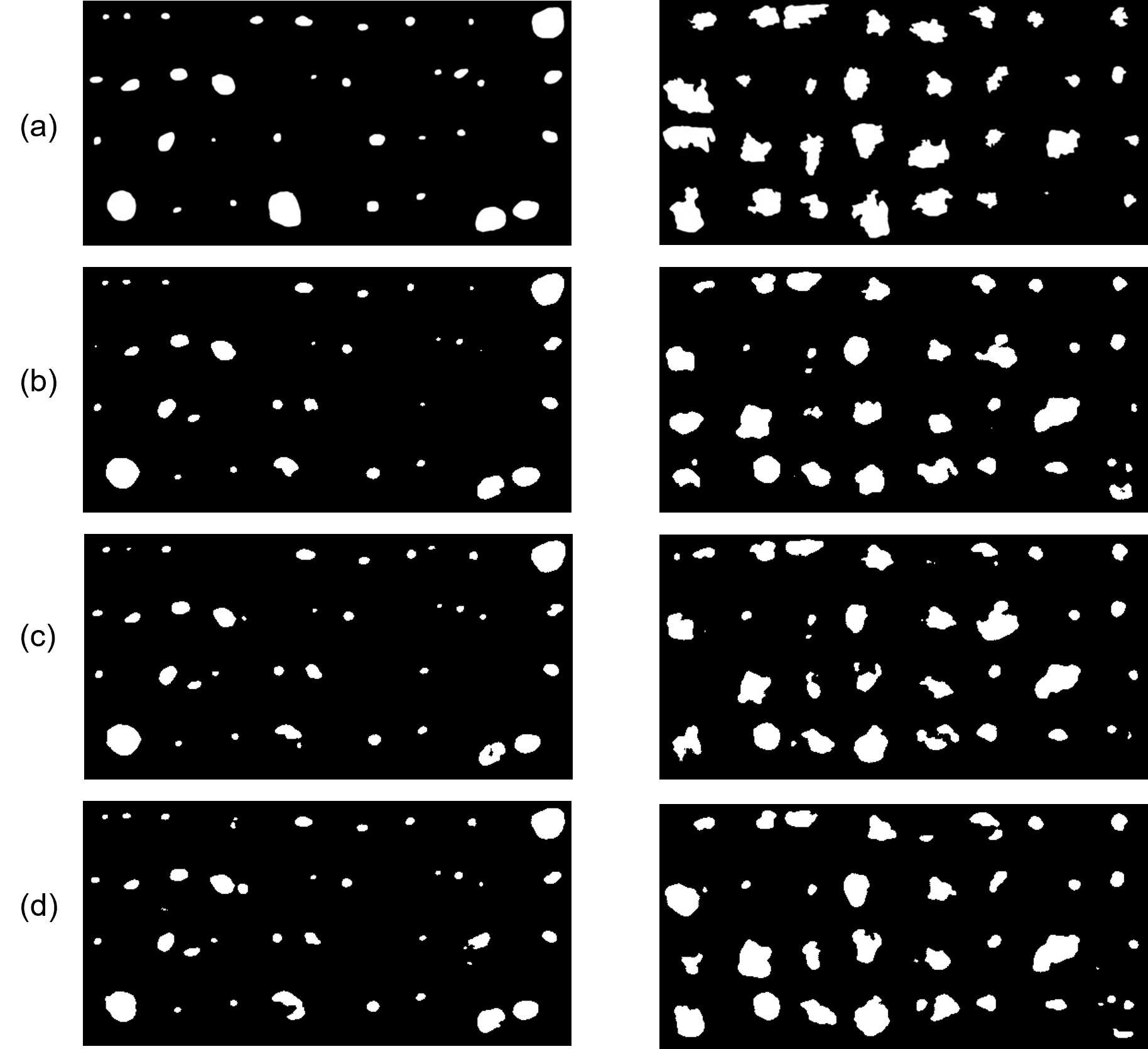}}
   \caption{Generated masks of the pre-trained ResNet-50 backbones, pre-trained and fine-tuned using $64\times64$ images. The first column shows the masks of benign tumours, and the second column shows the masks of malignant tumours. (a) Ground truth; (b) Supervised baseline; (c) SimSiam – BUS ($\bigcirc$) + mini-ImageNet; (d) SimSiam – Multi-organ ($\bigtriangleup$). 
}
%  \vspace{2.0cm}
\label{fig:resnet64_comp}
\end{figure}

\begin{figure}
  \centering
  \centerline{\includegraphics[width=9cm]{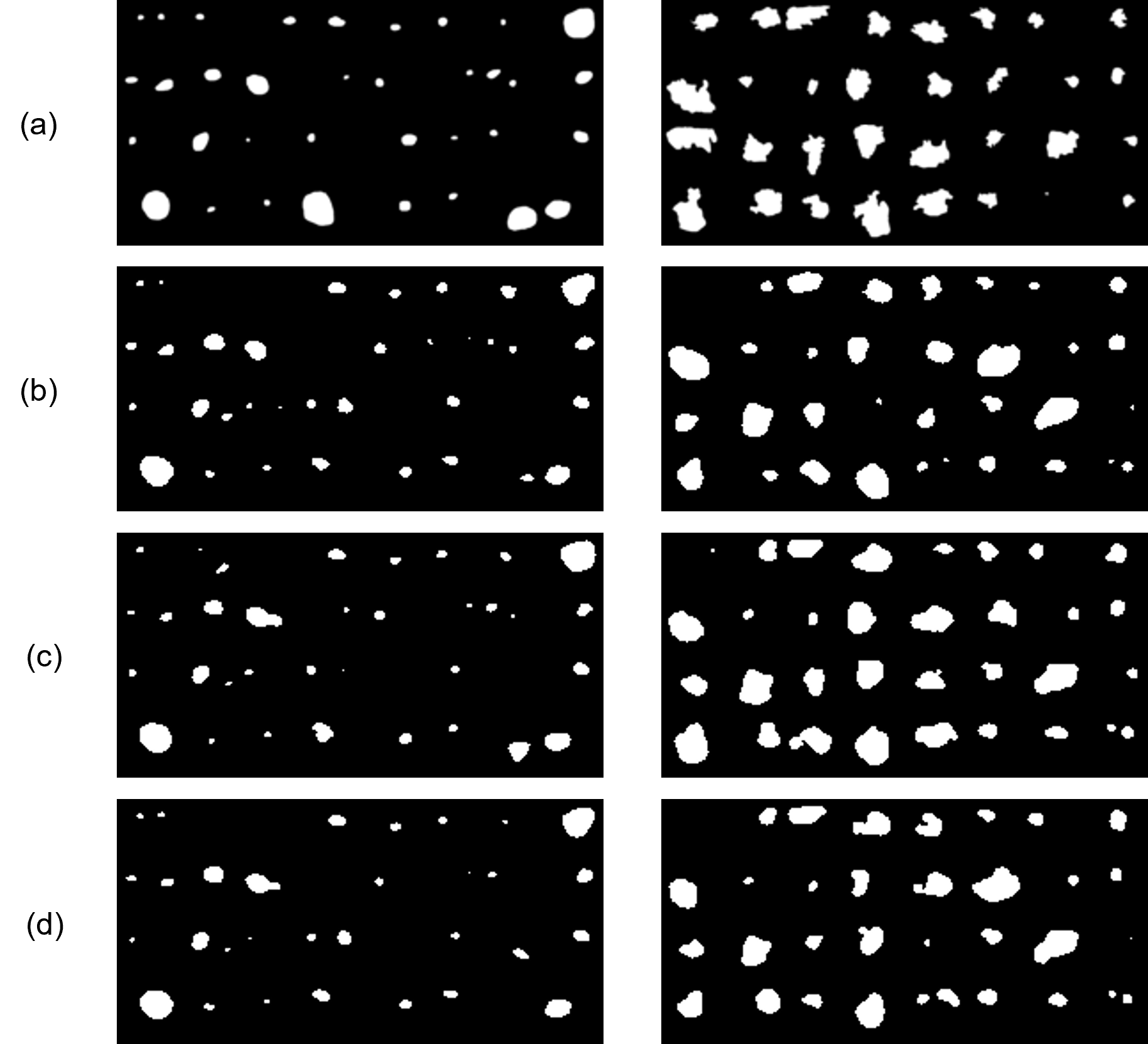}}
   \caption{Generated masks of the pre-trained U-Nets, pre-trained and fine-tuned using $32\times32$ images. The first column shows the masks of benign tumours, and the second column shows the masks of malignant tumours. (a) Ground truth; (b) Supervised baseline; (c) MoCo – BUS ($\bigcirc$) + CIFAR-10; (d) MoCo – Multi-organ ($\bigtriangleup$). 
}
%  \vspace{2.0cm}
\label{fig:unet32_comp}
\end{figure}

\begin{figure}
  \centering
  \centerline{\includegraphics[width=9cm]{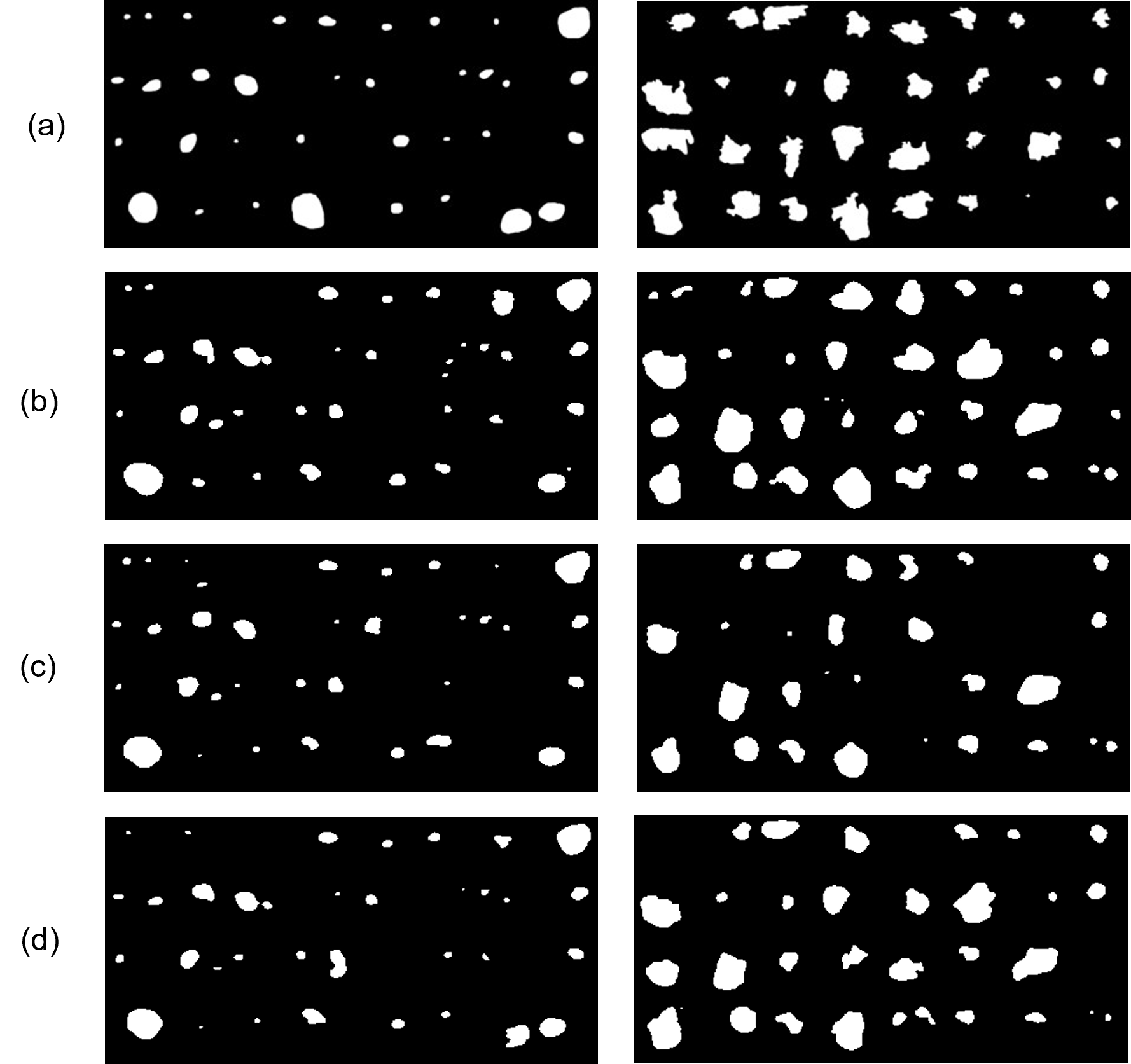}}
   \caption{Generated masks of the pre-trained U-Nets, pre-trained and fine-tuned using $50\times50$ images. The first column shows the masks of benign tumours, and the second column shows the masks of malignant tumours. (a) Ground truth; (b) Supervised baseline; (c) MoCo – BUS ($\bigcirc$); (d) MoCo – Multi-organ ($\bigtriangleup$).
}
%  \vspace{2.0cm}
\label{fig:unet50_comp}
\end{figure}

During the preparation of this work the author(s) used Grammarly in order to eliminate grammatical errors and improve word choice. After using this tool/service, the author(s) reviewed and edited the content as needed and take(s) full responsibility for the content of the publication.
\bibliographystyle{elsarticle-harv} 
\bibliography{cas-refs}





% \end{thebibliography}
\end{document}